\documentclass{article}
\usepackage[utf8]{inputenc}
\usepackage{authblk}

\usepackage{graphicx}
\usepackage{booktabs}
\usepackage{subfig}
\usepackage{amsmath}
\usepackage{url}
\usepackage{xcolor}
\usepackage{ulem}
\usepackage{cancel}
\usepackage{multirow}

\title{MedGCN: Medication Recommendation and Lab Test Imputation via Graph Convolutional Networks}
\date{}
\author[1]{Chengsheng Mao}
\author[1]{Liang Yao}
\author[1]{Yuan Luo}
\affil{Department of Preventive Medicine, Feinberg School of Medicine, Northwestern University, Chicago, IL, USA}

\begin{document}

\flushbottom
\maketitle
%
%
\thispagestyle{empty}

\begin{abstract}
Laboratory testing and medication prescription are two of the most important routines in daily clinical practice. Developing an artificial intelligence system that can automatically make lab test imputations and medication recommendations can save costs on potentially redundant lab tests and inform physicians of a more effective prescription. We present an intelligent medical system (named MedGCN) that can automatically recommend the patients' medications based on their incomplete lab tests, and can even accurately estimate the lab values that have not been taken. In our system, we integrate the complex relations between multiple types of medical entities with their inherent features in a heterogeneous graph. Then we model the graph to learn a distributed representation for each entity in the graph based on graph convolutional networks (GCN). By the propagation of graph convolutional networks, the entity representations can incorporate multiple types of medical information that can benefit multiple medical tasks. Moreover, we introduce a cross regularization strategy to reduce overfitting for multi-task training by the interaction between the multiple tasks. In this study, we construct a graph to associate 4 types of medical entities, i.e., patients, encounters, lab tests, and medications, and applied a graph neural network to learn node embeddings for medication recommendation and lab test imputation. we validate our MedGCN model on two real-world datasets: NMEDW and MIMIC-III. The experimental results on both datasets demonstrate that our model can outperform the state-of-the-art in both tasks. We believe that our innovative system can provide a promising and reliable way to assist physicians to make medication prescriptions and to save costs on potentially redundant lab tests. 
\end{abstract}


\section{Introduction}
With the increasing adoption of Electronic Health Records (EHRs), more and more researchers are working to mine clinical data to derive new clinical knowledge with a goal of enabling greater diagnostic precision, better personalized therapeutic recommendations in order to improve patient outcomes and more efficiently utilize health care resources ~\cite{winslow2012computational}. One of the key goals of precision medicine is the ability to suggest personalized therapies to patients based on their molecular and pathophysiologic profiles ~\cite{winslow2012computational}. Much work has been devoted to pharmacogenomics that investigates the link between patient molecular profiles and drug response ~\cite{karczewski2012pharmacogenomics}. However, the status quo is that patient genomics data is often limited while clinical phenotypic data is ubiquitously available. Thus there is great potential in linking patient pathophysiologic profiles to medication recommendation, a direction we termed as ``pharmacophenomics'' that is understudied yet will soon become imperative. This approach is particularly interesting for cancer treatment given the heterogeneous nature of the disease. In treating cancer patients, physicians typically prescribe medications based on their knowledge and experience. However, due to knowledge gaps or unintended biases, oftentimes these clinical decisions can be sub-optimal. 

On the other hand, the data quality issue often represents one of the major impediments of utilizing Electronic Health Record (EHR) data ~\cite{kohane2015ten}. Unlike experimental data that are collected per a research protocol, the primary role of clinical data is to help clinicians care for patients, so the procedures for its collection are often neither systematic nor on a regular schedule but rather guided by patient condition and clinical or administrative requirements.  Thus, many aspects of patients’ clinical states may be unmeasured, unrecorded and unknown in most patients at most time points.  While this missing data may be fully clinically appropriate, machine learning algorithms cannot directly accommodate missing data. Accordingly, missing clinical phenotypic data (e.g., laboratory test data) can hinder EHR knowledge discovery and medication recommendation efforts. 

Accurate lab test imputation can save costs of potentially redundant lab tests, and precise medication recommendations can assist physicians to formulate effective prescriptions. Developing an artificial intelligence (AI) medical system that can automatically and simultaneously perform the two medical tasks can help the medical institutions improve medical efficiency and reduce medical burden. However, most of the previous studies on intelligent medical systems performed the two tasks independently and ignored the interaction between them \cite{shang2019gamenet,wang2019neural,zhang2017leap,yu2020predict,waljee2013comparison,luo2016using,luo20173d}. Although most medication recommendation methods are based on the imputed lab tests, the error of lab test imputation is propagated to medication recommendation, increasing the error.

\begin{figure}[!t]
  \centering
  \includegraphics[width=\columnwidth,page=1]{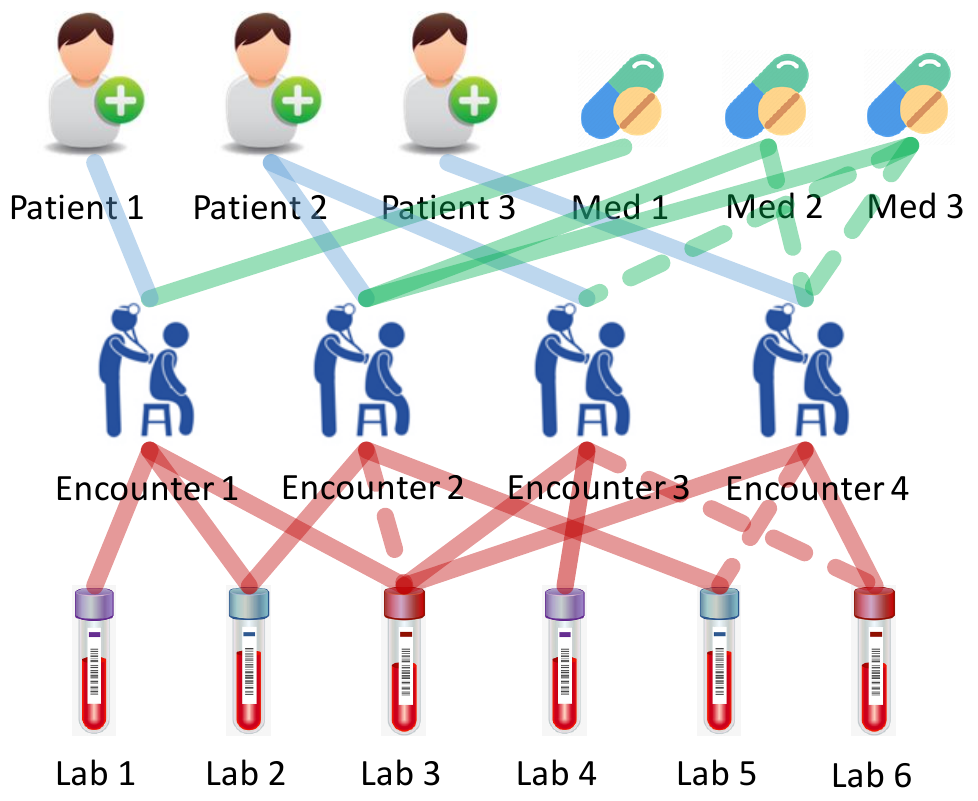}
  \caption{An example of MedGraph. The solid lines represent there are observed relations between the two objects; the dash lines represent the relation between the two objects are unknown. }
  \label{fig:example}
\end{figure}

However, as we all know, interactions can exist among various types of medical entities. For example, the lab test values usually indicate the health status of a patient, thus guide the medication recommendations. On the contrary, taking certain medications usually affects the lab test values. Since there are close and complex associations between medical entities, in this article, we incorporate the complex relations between multiple types of medical entities with their inherent features in a heterogeneous graph, named Medical Graph (MedGraph).  Figure \ref{fig:example} shows a MedGraph consisting of four types of nodes (i.e., encounters, patients, labs and medications), each of which could have their inherent features if they are available, e.g., demographic features for patients and the chemical composition for medications. Graph edges encode the relations between the medical entities, for example, a patient may have several encounters, an encounter may include some lab tests with certain values and some medication prescriptions. Figure \ref{fig:example} shows scenarios  where we may need to impute missing lab test results (e.g., Encounters 2, 3 and 4) and may need to recommend partial or a full list of medications (e.g., Encounters 3 and 4).

Recently, Graph Convolutional Networks (GCN) attracted wide attention for inducing informative representations of nodes and edges in graphs, and could be effective for tasks that could have rich relational information between different entities \cite{wu2020comprehensive,kipf2017semi,hamilton2017inductive,chen2018fastgcn,li2018classifying,yao2019graph,mao2019imagegcn}. However, there are two issues that must be effectively tackled before extending the original GCN to MedGraph: (1) the original GCN was designed for homogeneous graphs where all the nodes are of the same type and have the same feature names. (2) the original GCN could not handle missing values in the node features. In medical scenes, a MedGraph could be heterogeneous and have multiple types of nodes with different features, and there usually are many missing values in the feature matrix. In this article, based on the idea of GCN, we develop Medical Graph Convolutional Network (MedGCN) that is able to tackle the above issues for MedGraph and to learn informative distributed representations for nodes in MedGraph for better medication recommendation and lab test imputation. Our source code of MedGCN is available at \url{https://github.com/mocherson/MedGCN}.

The main contributions of this article are as follows. From the medical perspective, we provide an effective way of medication recommendation and lab test imputation based on EHR data. Moreover, in our experiments on real EHR data, our method MedGCN can significantly outperform the state-of-the-art methods for both medication recommendation and lab test imputation tasks. From the method perspective, we innovatively incorporate the complex associations between multiple medical entities into MedGraph, and develop MedGCN to learn the entity representations based on MedGraph for multiple medical tasks. From the AI perspective, (1) The developed MedGCN is able to perform multiple medical tasks within one model, including medication recommendations and lab test imputation. (2) MedGCN extends the GCN model to heterogeneous graphs and missing feature values in medical settings. (3) We introduce cross regularization, {a specific form of multi-task learning} to reduce overfitting for the training of MedGCN. (4) MedGCN is a general inductive model that could use the learned model to efficiently generate representations for new data. 

\section{Related Work}
In this section, we briefly review the literature on representation learning in the medical domain. More specifically, we focus on the tasks of medication recommendation and lab test imputation.

\subsection{Medical Representation Learning}
Representation learning aims to embed an object into a low-dimensional vector space while preserving the characteristics of the object as much as possible. Medical representation learning tries to generate vector representations for medical entities, e.g., patients and encounters. eNRBM \cite{tran2015learning} embedded patients into into a low-dimensional vector space with a modified restricted Boltzmann machine (RBM) for suicide risk stratification. DeepPatient \cite{miotto2016deep} derived patient representation from EHR data to facilitate clinical predictive modeling. Sun et al. \cite{sun2020disease} derived patient representations for disease prediction based on a medical concept graph and a patient record graph via graph neural networks. DoctorAI \cite{choi2016doctor} applied a gated recurrent unit (GRU)-based RNN model to embed patients for predicting the diagnosis codes in the next visit of patients. Med2Vec \cite{choi2016multi} learned the vector representations of medical codes and visits from large EHR datasets via a multi-layer perceptron (MLP) for clinical risk prediction. HORDE \cite{lee2020harmonized}, a unified graph representation learning framework, embedded patients' visits into a harmonized space for diagnosis code prediction and patient severity classification.  Decagon \cite{zitnik2018modeling} generated drug representations by a graph convolutional encoder to model polypharmacy side effects. HSGNN \cite{liu2020heterogeneous} generated visit representation via graph neural network for diagnosis prediction. MGATRx \cite{yella2021mgatrx} generated vector representations for drugs and diseases for drug repositioning. Most of the previous models focus on representations for only one type of medical entities (e.g., patients or encounters) and ignore the representation for entities of different types. Some models (e.g., Med2Vec, HORDE and HSGNN) can generate embeddings of multiple specific types of medical entities, but cannot be extended to more medical entities like lab test. Also, few of the previous works are designed for multiple tasks in one training. In our work, we incorporate multiple types of medical entities and their interactions into MedGraph which can be easily extended by adding new types of medical entities and their relations to exist entities. Moreover, the proposed MedGCN model can generate embeddings for all the nodes (medical entities) in MedGraph, thus feasible for multiple tasks involving the medical entities. 

\subsection{Graph Representation Learning on EHR}
Graph representation learning tries to learn a vector representation for part or the whole of a graph for certain graph learning tasks, e.g., learning node representation for node classification. As a simple and effective graph learning model, GCN is increasingly applied to the medical domain for EHR mining and analysis. Since EHR usually contains multiple medical entities and multiple relations, the graphs derived from EHR are usually heterogeneous, thus, the original GCN cannot be directly applied to heterogeneous graphs. We found a number of methods in the literature applied GCN to graphs from EHR. HORDE \cite{lee2020harmonized} constructed a multi-modal EHR graph caontaining at least 3 types of nodes, since node features are not available, HORDE ignored the node types and applied GCN to the EHR graph to learn node representations for diagnosis code prediction and patient severity classification. Sun et al. \cite{sun2020disease} constructed a medical concept graph and a patient record graph for disease prediction; both graphs are a bipartite graph with two types of nodes. They ignored the node types in the propagation rule, but applied a projection weight to align all node embeddings onto the same space. SMGCN \cite{jin2020syndrome} construct multiple graphs (i.e., the symptom-herb bipartite graph, symptom-symptom graph, and herb-herb graph) for herb recommendation; it used a bipar-GCN for information propagation on the symptom-herb bipartite graph. In our work, we decompose a heterogeneous graph into multiple bipartite graphs or homogeneous graphs where the message passing to a certain node is from only one type of nodes, then aggregate all the messages the node received to generate the node embedding. From this view, the bipar-GCN is a special case of our model for bipartite graphs.

MGATRx \cite{yella2021mgatrx} constructed a graph with 6 node types from different EHR data, and then applied attention mechanism to extend the multi-view sum pooling for drug repositioning. MGATRx and our model MedGCN are both based on GCN for heterogeneous graphs. But there are server major differences between our work. (1) We have different medical tasks, MGATRx is for drug repositioning. MedGCN is for medication recommendation and lab task imputation. (2) The constructed graphs are different. The graph in MGATRx has 6 types of nodes (i.e., drug, disease, target, substructure, side effect, Mesh), while our graph contains 4 types of nodes (patients, medications, labs, encounters). The only common node type is drug/medication. (3) Our MedGCN has a wide range of applicability, MGATRx paper also applied MedGCN to its graph for drug repositioning and achieved good performance, though not better than MGATRx. But it is not straightforward to apply MGATRx to weighted heterogeneous graphs which is exactly our case.

\subsection{Medication Recommendation}
Following the categorization from ~\cite{shang2019gamenet}, there are mainly two types of deep learning methods designed for medication recommendation: instance-based methods and Longitudinal methods. instance based methods perform recommendations ignoring the longitudinal patient history, e.g.  LEAP ~\cite{zhang2017leap} formulated treatment recommendation as a reinforcement learning problem to predict combination of medications given patient’s diagnoses. SMGCN \cite{jin2020syndrome} applied GCN to multiple graphs to describe the complex
relations between symptoms and herbs for herb recommendation. Longitudinal methods leverage the temporal dependencies within longitudinal patient history to predict future medication, e.g., DMNC ~\cite{le2018dual} considers the interactions between two sequences of drug prescription and disease progression in the model. GAMENet ~\cite{shang2019gamenet} models longitudinal patient records in order to provide safe medication recommendation. G-BERT \cite{shang2019pre} combined GNN and transformer models to generate visit representations for medication recommendation. However, these work focused on one task while not considering how to integrate with other tasks. Moreover, most of the previous work for medication recommendation took only the diagnosis code as an input and were unable to directly accommodate lab tests as input. In this work, to release the workload of physicians in diagnostic reasoning, our model makes the recommendation based on a graph containing lab tests and patients rather than the diagnosis codes.

\subsection{Lab Test Imputation}
Clinical data often contain missing values for test results. Imputation uses available data and relationships contained within it to predict point or interval estimates for missing values. Numerous imputation algorithms are available ~\cite{waljee2013comparison,buuren2010mice,luo2016using,stekhoven2011missforest}, many of these are designed for cross-sectional imputation (measurements at the same time point). Recent imputation studies have attempted to model the time dimension ~\cite{he2011functional,luo20173d,kliethermes2014bayesian}, but they generally consider all time points to occur within the same patient encounters where the temporal correlation between these time points are strong enough to contribute to imputation accuracy. In our work, the lab test imputation is based on a MedGraph that incorporates multiple types of medical entities and their relations.

\section{Methods}
\subsection{Problem Formulation}

Since an encounter may include multiple medications, we model medication recommendation as a multi-label classification problem. For each medication, the model should output the recommendation probability. If an informative representation can be learned for each encounter, this problem can be tackled by many traditional machine learning algorithms. Thus the problem is how to effectively integrate the lab information and patient information into the representations of encounters. 

Missing lab values imputation is another challenge in encounter representation learning. Using an encounter-by-lab matrix, the imputation problem can be formulated as a matrix completion problem. Common imputation methods such as mean or median imputation will overlook the correlation between different columns. In the medical setting, the concurrent information such as patients' baselines and medications are also useful for lab test imputation. We integrate all this information in the encounter representations to help impute the missing labs.  

Both the two tasks require informative encounter representations. Thus, we can conduct the two tasks with one model MedGCN where each node can get information from its neighbors in a layer, and eventually, the encounter nodes could learn informative representations that can be used for both medication recommendation and lab test imputation. In addition, we can train the model with a cross regularization strategy by which the loss of one task can be regarded as a regularization item for the other task.

\subsection{MedGraph} \label{sec:medgraph}
We define MedGraph as a specialized graph $G=(\mathcal{V},\mathcal{E})$ where the nodes $\mathcal{V}$ consist of medical entities and the edges $\mathcal{E}$ consist of the relations between the medical entities. MedGraph is a heterogeneous graph where the nodes $\mathcal{V}$ consist of multiple types of medical entities. Our goal is to learn a vector representation for each node in the graph.

Since GCN cannot directly handle missing values in node features, this leads to an issue for GCN to consider labs as features of encounters due to many missing values in labs. Fortunately, a graph can accept an empty edge between two nodes, thus, we represent the labs as nodes of type ``lab" that connect to the encounter nodes, i.e., we construct a bipartite subgraph between labs and encounters as a part of MedGraph.  

\begin{figure*}[!t]
  \centering
  \includegraphics[width=\textwidth,page=1]{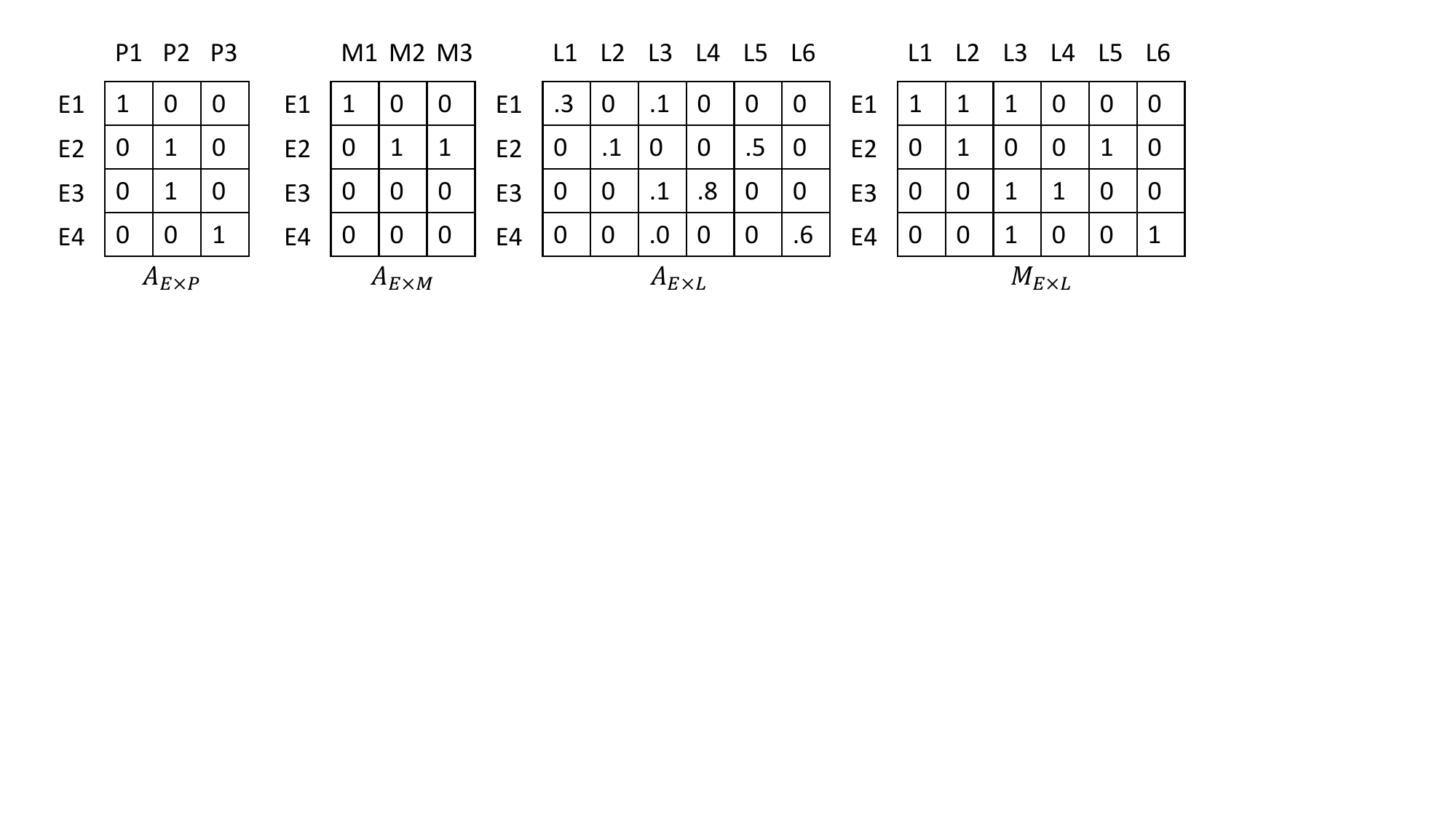}
  \caption{Adjacency matrices corresponding to MedGraph, $A_{E\times P}$: adjacency matrix between encounters and patients; $A_{E\times M}$: adjacency matrix between encounters and medications; $A_{E\times L}$: adjacency matrix between encounters and labs; $M_{E\times L}$: mask matrix between encounters and labs. Abbreviations: P - patient; E - encounter; M - medication; L - lab.}
  \label{fig:adjmats}
\end{figure*}

Figure \ref{fig:example} illustrates an example of our MedGraph where the nodes consist of four types of medical entities, i.e., Encounters, Patients, Labs, Medications, each node could have inherent features within it. The relations between two types of nodes correspond to an adjacency matrix. We defined the relations between two types of nodes as follows, and the example adjacency matrices corresponding to the MedGraph in Figure \ref{fig:example} are shown in Figure \ref{fig:adjmats}.

\begin{itemize}
\item The relations between encounters and patients are defined in matrix $A_{E\times P}$ in Figure \ref{fig:adjmats}. If an encounter belongs to a patient, there is an edge between the two nodes, the corresponding position of the adjacency matrix $A_{E\times P}$ is set to 1, otherwise, 0. Since one encounter must belong to one and only one patient, there must be only one ``1" in each row of the matrix $A_{E\times P}$.

\item The relations between encounters and labs are defined in matrix $A_{E\times L}$ in Figure \ref{fig:adjmats}. If an encounter includes a lab test, there is a weighted edge between the two nodes, the corresponding position of the adjacency matrix $A_{E\times L}$ is set to the test value scaled to $[0,1]$ by min-max normalization as Eq. \ref{eq:norm}, otherwise, 0. To discriminate the true test values ``0'' and the missing values, we introduce the mask matrix $M_{E\times L}$ with the same shape as $A_{E\times L}$, where the entries corresponding to true test values are set to 1, and the entries corresponding to missing values are set to 0. For example, if the L1 test value of E2 is 0, the (E2, L1) position is 0 in the $A_{E\times L}$ matrix, but is 1 in the mask $M_{E\times L}$ matrix (Figure \ref{fig:adjmats}).

\item The relations between encounters and medications are defined in matrix $A_{E\times M}$ in Figure \ref{fig:adjmats}. If an encounter is ordered a medication, there is an edge between the two nodes, the corresponding position of the adjacency matrix $A_{E\times M}$ is set to 1.  

\item The connections between nodes of the same type are all self-connections, the corresponding adjacency matrices are all set as the identity matrix. 
\end{itemize}

In practice, there could be more nodes, more types of medical entities, and more relations that can be incorporated into MedGraph. For example, if medication similarities are known, we could construct a $N_M\times N_M$ matrix to represent the relations between medications. However, such similarity information often requires external knowledge sources that are not always available and may be subjective, thus we only consider the above 4 types of relations in this study. Note that we did not use diagnosis codes for two reasons. Diagnosis codes often are recorded for billing purposes and may not reflect patients' true pathology. Moreover, we want to make MedGCN more practically useful by not asking physicians to do the heavy lifting in diagnostic reasoning. Nevertheless, the diagnosis codes can be easily added to the MedGraph as a new node type for medication recommendation and lab test imputation.  

\subsection{Graph Convolutional Networks} \label{sec:model}

Recently, graph convolutional networks (GCN) attracted wide attention for inducing informative representations of nodes and edges, and could be effective for tasks that could have rich relational information between different entities \cite{kipf2017semi,hamilton2017inductive,zhang2018multi,chen2018fastgcn,li2018classifying,yao2019graph,hsieh2020drug}. 

\subsubsection{General GCN}
GCN learns node representations based on the node features and their connections with the following propagation rule for an undirected graph \cite{kipf2017semi}:
\begin{equation}\label{eq:GCN}
H^{(k+1)} = \phi (\tilde{D}^{-\frac{1}{2}} \tilde{A} \tilde{D}^{-\frac{1}{2}} H^{(k)} W^{(k)})
\end{equation}
where $\tilde{A}=A+I$ is the adjacency matrix with added self-connection, $\tilde{D}$ is a diagonal matrix with $\tilde{D}_{ii}=\sum_j{\tilde{A}_{ij}}$, $H^{(k)}$ and $W^{(k)}$ are the node representation matrix and the trainable parameter matrix for the $k$th layer, respectively; $H^{(0)}$ can be regarded as the original feature matrix, $\phi(\cdot)$ is the activation function. 

The propagation rule of GCN can be interpreted as the Laplacian smoothing \cite{li2018deeper}, i.e., the new representation of a node is computed as the weighted average of itself and its neighbors, followed by a linear transformation. Further, GCN can be generalized to a graph neural network where each node can get and integrate messages from its neighborhood to update its representation \cite{morris2019weisfeiler,hamilton2017representation}, i.e., a node $v$'s representation is updated by
\begin{equation}\label{eq:generalGNN}
 H^{(k+1)}(v) =  
f_{1}^{W_1^{(k)}} \left(H^{(k)}(v) , f_{2}^{W_2^{(k)}}(\{H^{(k)}(w) | w\in N(v)\}) \right)
\end{equation}
where $N(v)$ is the neighborhood of node $v$, $f_{2}^{W_2}$ aggregates over the set of neighborhood features and $f_{1}^{W_1}$ merges the node and its neighborhood's representations. Both $f_{1}^{W_1}$ and $f_{2}^{W_2}$ should be arbitrary differentiable functions. Since a neighborhood is a set that is permutation-invariant, the aggregation function $f_{2}^{W_2}$ should also be permutation-invariant. If $f_{1}^{W_1}$ and $f_{2}^{W_2}$ are sum operations, Eq. \ref{eq:generalGNN} is simplified as Eq. \ref{eq:generalGCN} \cite{schlichtkrull2018modeling}.
\begin{equation}\label{eq:generalGCN}
H^{(k+1)}(v) = \phi \left(\sum_{w\in N(v)\cup\{v\}} H^{(k)}(w)\cdot W^{(k)}  \right)
\end{equation}

Eq. \ref{eq:generalGCN} is designed for a homogeneous graph where all the samples in $\tilde{N}(v) = N(v)\cup\{v\}$ share the same linear transformation $W^{(k)}$. It can be shown that Eq. \ref{eq:generalGCN} is equivalent to Eq. \ref{eq:GCN} except for the normalization coefficients. 

\subsubsection{Heterogeneous GCN}
Since the multiple types of nodes imply multiple types of edges in any heterogeneous graphs, the key problem of applying GCN to heterogeneous graphs is how to discriminate different types of edges in a heterogeneous graph when doing the propagation. In our method, we decompose the heterogeneous graph into multiple subgraphs with each subgraph has only one type of edge which can be represented by a single adjacency matrix. In each GCN layer, the representations of each node in all the subgraphs are aggregated as the node representation. Based on this idea, Eq.~\ref{eq:generalGCN} can be extended to multiple types of edges as Eq. \ref{eq:heteroGCN}. 
\begin{equation}\label{eq:heteroGCN}
H^{(k+1)}(v) = \phi \left(\sum_{i=1}^n \sum_{w\in \tilde{N}_i(v)} H^{(k)}(w)\cdot W_i^{(k)}  \right)
\end{equation}
where $\tilde{N}_i(v)$ is the neighborhood of $v$ ($v$ included) in terms of $i$th edge type. Eq. \ref{eq:heteroGCN} can also be written in matrix form as

\begin{equation}\label{eq:matheteroGCN}
H_i^{(k+1)} = \phi \left(\sum_{j=1}^n A_{ij} \cdot H_j^{(k)}\cdot W_{ij}^{(k)}  \right)
\end{equation}
where $H_i^{(k)}$ is the presentation of the $i$th type nodes in the $k$th layer, $A_{ij}$ is the adjacency matrix between nodes type $i$ and $j$, $W_{ij}^{(k)}$ is the learnable linear transformation matrix indicating how type $j$ nodes contribute to type $i$ nodes' representation in layer $k$.\footnote{This can be scaled to a multi-graph where multiple types of edges between two nodes exist by aggregating multiple items of the $A\cdot H\cdot W$.}

Many existing works have extended GCN to heterogeneous graphs like (\cite{schlichtkrull2018modeling,wang2019heterogeneous,zitnik2018modeling}), and quite a few works represented EHR as graphs and employed graph-based method for certain medical tasks \cite{liu2020heterogeneous,yella2021mgatrx,shang2019pre}. However, the MedGraph involved in our work is a specific heterogeneous graph that can incorporate weighted edges, and different from general weighted graphs, edge with weight $\le 0$ can exist, because a lab value can be 0 or negative (e.g., base excess). Although many existing GCN models can deal with a general heterogeneous graph, few of the models can be applied to MedGraph directly. The significance of our work is to introduce to GCN model for a more specific heterogeneous graph like MedGraph.

\subsection{MedGCN}

\begin{figure*}[!t]
  \centering
  \includegraphics[width=\textwidth,page=1]{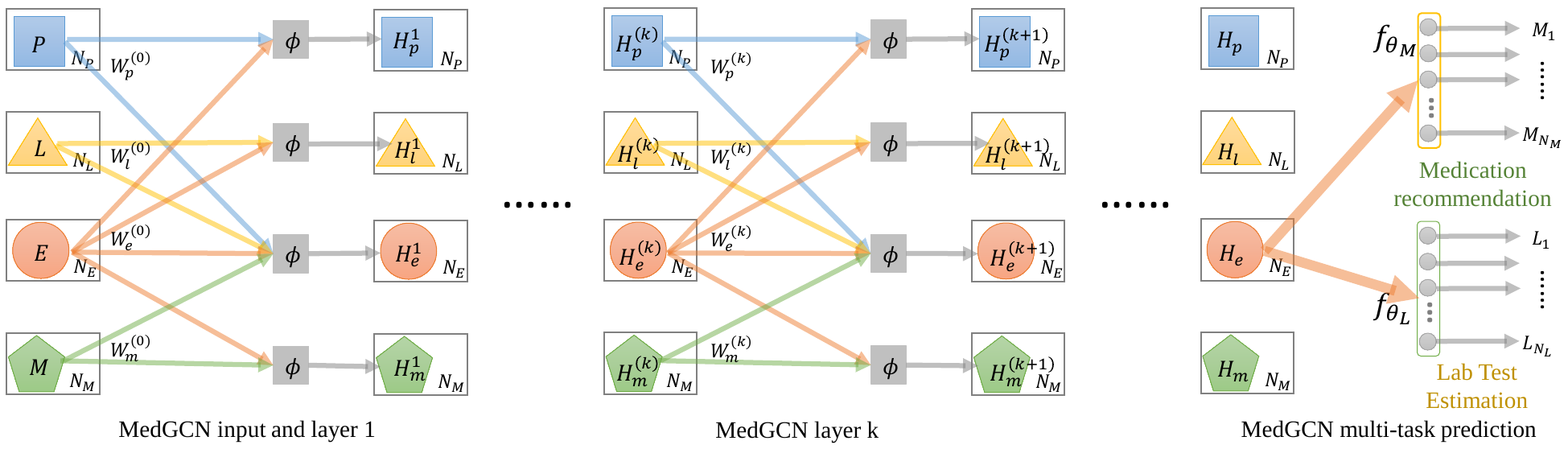}
  \caption{Architecture of MedGCN. Shape and color of the nodes indicate different types of nodes. Here we use plate notation where the numbers in the corners of plates ($N_P, N_L, N_E, N_M$) indicate repetitions of the same type of nodes in the MedGraph. Edges with the same color possess shared weights (parameter sharing).  P (patient), E (encounter), M (medication), L (lab) are MedGCN input and $H_p^{(k)}, H_e^{(k)}, H_m^{(k)}, H_l^{(k)}$ are hidden representations of corresponding nodes at layer $k$.}
  \label{fig:propagation}
\end{figure*}

Eq. \ref{eq:matheteroGCN} is a general formula for GCN propagation in a heterogeneous graph. Applying the propagation rule Eq.~\ref{eq:matheteroGCN} to our MedGraph defined in Section \ref{sec:medgraph}, we design our MedGCN architecture as shown in Figure \ref{fig:propagation}; and the propagation for different types of entities are  
\begin{equation}
\begin{aligned}
    H_e^{(k+1)} &= \scriptstyle \phi \left(A_{E\times P}\cdot H_p^{(k)}\cdot W_p^{(k)} 
                + A_{E\times L}\cdot H_l^{(k)}\cdot W_l^{(k)} 
                + A_{E\times M}\cdot H_m^{(k)}\cdot W_m^{(k)} 
                + H_e^{(k)}\cdot W_e^{(k)} 
                \right)   \\
    H_p^{(k+1)} &= \scriptstyle \phi \left(A_{P\times E}\cdot H_e^{(k)}\cdot W_e^{(k)} 
                + H_p^{(k)}\cdot W_p^{(k)} 
                \right)  \\
    H_l^{(k+1)} &= \scriptstyle  \phi \left(A_{L\times E}\cdot H_e^{(k)}\cdot W_e^{(k)} 
                + H_l^{(k)}\cdot W_l^{(k)} 
                \right)  \\
    H_m^{(k+1)} &= \scriptstyle  \phi \left(A_{m\times E}\cdot H_e^{(k)}\cdot W_e^{(k)} 
                + H_m^{(k)}\cdot W_m^{(k)} 
                \right) 
\end{aligned}   
\label{eq:medgcn}
\end{equation}
where $A_{E\times P}$, $A_{E\times M}$ and $A_{E\times L}$ are defined as in Section \ref{sec:medgraph}, their transposes are denoted as $A_{P\times E}$, $A_{L\times E}$ and $A_{M\times E}$, respectively, $H_p^{(k)}$, $H_e^{(k)}$, $H_m^{(k)}$, and $H_l^{(k)}$ are the representations of the corresponding type of nodes at layer $k$. We make all the transformations from the same type of nodes share parameters to reduce model complexity,  $W_p^{(k)}$, $W_e^{(k)}$, $W_m^{(k)}$, and $W_l^{(k)}$ are the linear transformation matrix from the corresponding type of nodes. Since the adjacency matrix between the same node type is identical, we omit it in Eq. \ref{eq:medgcn}. 

Because encounters have connections to patients, labs and medications, an encounter node updates its representation based on information from neighbors of these types and itself. Information propagation on other nodes is similar. After propagation in every layer, each node representation incorporates information from its 1-hop neighbors. Eventually, in the last layer (assume $k$th layer), each node learns an informative distributed representation based on its $k$-hop neighborhood information. The representations are then inputted to two different fully-connected neural networks $f_{\theta_M}$ and $f_{\theta_L}$ followed by the sigmoid activation for medication recommendation and lab test imputation, respectively, i.e., 
\begin{equation}\label{eq:medlaboutput}
\begin{aligned}
P & = sigmoid(f_{\theta_M}(H_e))  \\
V & = sigmoid(f_{\theta_L}(H_e))
\end{aligned}
\end{equation}
where $H_e$ is the final encounter representations, the output $P$ is a matrix of size $N_E \times N_M$ with the entry $p_{ij}$ is the recommendation probability of medication $j$ for encounter $i$; the output $V$ is a matrix of size $N_E \times N_L$ with the entry $v_{ij}$ is the normalized estimated value of lab $j$ for encounter $i$. Here, $N_E$, $N_M$, and $N_L$ is the number of encounters, medications and labs, respectively. 


\subsection{Model training} \label{sec:training}
\textbf{Loss function.}
For the medication recommendation task, the true label is the adjacency matrix $A_{E\times M}$. Since $A_{E\times M}$ is very sparse, to favor the learning of positive classes, we give a weight $N_n/N_p$  to the positive classes, where $N_n$ and $N_p$ are the counts of `0's and `1's in $A_{E\times M}$, respectively. Similar with the previous work \cite{mao2018deep,wang2017chestxray}, The binary cross entropy loss for medication recommendation is defined as 
\begin{equation}\label{eq:Mloss}
\mathcal{L}_M(P,A) = -\frac{1}{N_E}\sum_i^{N_E}\sum_j^{N_M}  \left( \frac{N_n}{N_p} a_{ij}\log{p_{ij}} +(1-a_{ij})\log{(1-p_{ij})} \right)
\end{equation}
where $a_{ij}$ and $p_{ij}$ are the values at position $(i,j)$ in matrix $A$ and $P$ respectively.

For the lab test imputation task, the adjacency matrix $A_{E\times L}$ contains the true values as well as some `0's for the missing values. We use the mask matrix $M_{E\times L}$ to screen the true targets for training. Since the lab value are continuous, we use the mean square error loss for lab test imputation as 
\begin{equation}\label{eq:Lloss}
\mathcal{L}_L(V,A) = -\frac{1}{N_E}\sum_i^{N_E}\sum_j^{N_L} m_{ij}(v_{ij}-a_{ij})^2  
\end{equation}
where $a_{ij}$, $v_{ij}$ and $m_{ij}$ are the values at position $(i,j)$ in matrix $A$, $V$ and $M$ respectively.

To learn a MedGCN model that can perform the two tasks, we use the following loss
\begin{equation}\label{eq:loss}
\mathcal{L}=\mathcal{L}_M(P,A_{E\times M})+\lambda\mathcal{L}_L(V,A_{E\times L})
\end{equation}
where $\lambda$ is a regularization parameter used to adjust the proportion of the two losses. 

\textbf{Cross regularization.} 
Generally, for the medication recommendation task, we only need to minimize $\mathcal{L}_M$ loss, and for lab test imputation, we only need $\mathcal{L}_L$ loss. {For both tasks, we need to minimize both $\mathcal{L}_M$ and $\mathcal{L}_L$. Thus, we achieve the loss function Eq. \ref{eq:loss} where $\mathcal{L}_L$ can be seen as a regularization item for the medication recommendation task, and $\mathcal{L}_M$ can be seen as a regularization item for the lab imputation task. We call this strategy cross regularization between multiple tasks. Cross regularization is a specific form of multi-task learning for our scenario to reduce overfitting. Generally, multi-task learning has a major task and one or more auxiliary tasks, the auxiliary tasks serve the major task in the form of regularization. However, in our scenario, the two tasks medication recommendation and lab test imputation are both major tasks. They server as a regularization item for each other.} Beneficially, cross regularization could make the learned representations more informative, because they would carry information from other related tasks.

\textbf{Inductive learning.}
The original GCN is considered transductive, i.e., it requires that all nodes including test nodes in the graph are present during training, only the label to predict is unknown. This will make the model not be able to generalize to unseen nodes. There were some inductive versions of GCN that can generalize to unseen new samples \cite{hamilton2017inductive,chen2018fastgcn}. Our MedGCN can also be implemented inductively based on Eq. \ref{eq:heteroGCN}. For a trained model, all its node embeddings $H_*^{(k)}$ and the learned parameters $W_*^{(k)}$ are known, if a new test sample $v$ and its connections to the graph are known, the embedding of neighbors of $v$ could be retrieved to compute the embedding of $v$ by Eq. \ref{eq:heteroGCN} without retraining the model.

When a new encounter comes, $H_e^{(0)}$ has one more row with the original feature of the new encounter and $A_{E\times P}$,$A_{E\times L}$,$A_{E\times M}$ all have one more row corresponding to the connection of the new encounter to the available patients, labs and medications, respectively. Since the transformation matrix $W_e^{(0)}$ has been already trained, $H_e^{(0)} \cdot W_e^{(0)}$ has one more row corresponding to the new encounter. Similarly, $A_{E\times P}\cdot H_p^{(0)} \cdot W_p^{(0)}$, $A_{E\times L}\cdot H_l^{(0)} \cdot W_l^{(0)}$  and $A_{E\times M}\cdot H_m^{(0)} \cdot W_m^{(0)}$ all have one more row corresponding to the new encounter, thus $H_e^{(1)}$ has one more row for the new encounter by Eq. \ref{eq:medgcn}.  Finally, $H_e^{(k)}$ has one more row corresponding to the embedding of the new encounter. 

To perform medication recommendation and lab test imputation for a new encounter $e$ with certain lab test values unknown, we first add the encounter $e$ to MedGraph based on the known information according to the description in Section \ref{sec:medgraph} and then perform the propagation rules (i.e., Eq. \ref{eq:medgcn}) of the trained MedGCN on the updated MedGraph. Eventually, we get the final representation $h_e$ for the encounter. After processed by the fully-connected layers $f_{\theta_M}$ and $f_{\theta_L}$, we get the output medication probability $p = sigmoid(f_{\theta_M}(h_e))$, and the estimated lab test value $v = sigmoid(f_{\theta_L}(h_e))$, respectively, where $p$ is a vector of $N_M$ components corresponding to the recommendation probabilities of the $N_M$ medications, and $v$ is a vector of $N_L$ components corresponding to the estimated values of the $N_L$ lab tests.

\section{Experiments}
Our method can be applied to a wide variety of MedGraphs and medical tasks. Here, we validated it on {two real-world EHR datasets, NMEDW and MIMIC-III, for medication recommendation and lab test imputation. NMEDW is a dataset of patients with lung cancer from Enterprise Data Warehouse (EDW) of Northwestern Medicine \cite{starren2015enabling}. MIMIC-III \cite{johnson2016mimic} is a public EHR dataset comprising information relating to patients admitted to critical care units over 11 years between 2001 and 2012.} 

\subsection{Data preparation} \label{sec:datasource}
{For NMEDW dataset,} we prepared the dataset by the following steps. (1) We identified a cohort of patients with lung cancer using ICD-9 codes 162.* and 231.2 between 1/1/2004 and 12/31/2015, 5017 patients and their 112510 encounters were identified. (2) We considered the list of cancer drugs approved by National Cancer Institute\footnote{\url{https://www.cancer.gov/about-cancer/treatment/drugs}} for cancers or conditions related to cancer for medication recommendation, 518 cancer drugs were considered. (3) We excluded the encounters that had no cancer drug medication records, 9638 encounters corresponding to 1277 patients were left. (4) We excluded the labs that only a few ($<1\%$) encounters took and the labs that were constant for all the encounters, 197 labs were identified. (5) We excluded the encounters that did not take any of the 197 labs, getting 1260 encounters corresponding to 865 patients. (6) We excluded the cancer drugs that were never prescribed to any of the remaining 1260 encounters, getting 57 cancer drug medications. Eventually, the NMEDW dataset consists of 1260 encounters for 865 patients, they have taken 197 labs and 57 medications related to lung cancer. 

{For MIMIC-III dataset, we prepared the dataset by the following steps. (1) The same 518 cancer drugs are also considered on NMEDW dataset. (2) We excluded the encounters (i.e., visits in MIMIC-III) that had no cancer drug medication records, 18223 encounters corresponding to 15182 patients were left. (3) We excluded the labs that only a few ($<1\%$) encounters took and the labs that were constant for all the encounters, 219 labs were identified. (4) We excluded the encounters that did not take any of the 219 labs, getting 18190 encounters corresponding to 15153 patients. (5) We excluded the cancer drugs that were never prescribed to any of the remaining 18190 encounters, getting 117 cancer drug medications. Eventually, the MIMIC-III dataset consists of 18190 encounters for 15153 patients, they have taken 219 labs and 117 medications.} 

{For both datasets,} each lab value is scaled to [0,1] by min-max normalization as \begin{equation}
v=\frac{v_{org}-v_{min}}{v_{max}-v_{min}}  \label{eq:norm}  
\end{equation}
where $v_{org}, v_{min}, v_{max}$ are the original value, the maximum value and the minimum value of the corresponding lab test, respectively, and $v$ is the normalized value. Based on the final dataset, we can construct a MedGraph similar to Figure \ref{fig:example} as well as its adjacency matrix according to the criterion described in Section \ref{sec:medgraph}, but we have more nodes and edges. The MedGraph information constructed on {NMEDW and MIMIC-III} datasets is summarized in Table \ref{tab:graph} {and \ref{tab:graph_mimic}, respectively.}

\begin{table}[htbp]
\caption{Summary information of {NMEDW} dataset and the corresponding MedGraph. E=encounters; P=patients; L=labs; M=medications}
  \centering
    \#E: 1260; \#P: 865; \#L: 197, \#M: 57 \\
    \begin{tabular}{lllrr}
    \toprule
    Matrix & Size & {Edges} & {Sparsity}   & Values \\
    \midrule
    $A_{E\times P}$ & 1260 $\times$ 865 & 1260  & 99.88\% & binary: 0, 1 \\
    $A_{E\times L}$ & 1260 $\times$ 197 & 43806 & 82.35\% & continuous: 0--1 \\
    $A_{E\times M}$ & 1260 $\times$ 57 & 2475  & 96.55\% &  binary: 0, 1 \\
    \bottomrule
    \end{tabular}   
 
  \label{tab:graph}%
\end{table}%

\begin{table}[htbp]
\caption{{Summary information of MIMIC-III dataset and the corresponding MedGraph. E=encounters; P=patients; L=labs; M=medications} }
  \centering
    \#E: 18190; \#P: 15153; \#L: 219, \#M: 117 \\
    \begin{tabular}{lllrr}
    \toprule
    Matrix & Size & {Edges} & {Sparsity}   & Values \\
    \midrule
    $A_{E\times P}$ & 18190 $\times$ 15153 & 18190  & 99.99\% & binary: 0, 1 \\
    $A_{E\times L}$ & 18190 $\times$ 219 & 1029964 & 68.96\% & continuous: 0--1 \\
    $A_{E\times M}$ & 18190 $\times$ 117 & 23395  & 98.68\% &  binary: 0, 1 \\
    \bottomrule
    \end{tabular}   
  \label{tab:graph_mimic}%
\end{table}%

{For both datasets and both tasks,} we randomly split the dataset into a train-val set and a test set with proportion 8:2, the train-val set is further randomly split into training set and validation set by 9:1. For the medication recommendation task, the patients are split into training, validation and test. To be practical, a model should use previous encounter information to make recommendations for later encounters, thus in our setting, the last encounters of the test patients and the validation patients constitute the test encounter set and the validation encounter set, respectively; all other encounters are used for training. Since our datasets contain many patients with only one encounter (specifically, 141 out of 173 test patients {for NMEDW and 2628 out of 3031 test patients for MIMIC-III}), the test performance can represent the generalization capability of a model to new patients. In the training process, we remove all edges from the test and validation encounters to any medications in MedGraph. For inductive MedGCN, we remove all testing encounters and their connections from MedGraph when training. For lab test imputation task, the edges between encounters and labs are split into training, validation and test. In the training process, we remove all validation and testing edges corresponding to the missing values in MedGraph.  For both tasks, in each training epoch we evaluate the performance on the validation set. The best model for the validation set is saved for evaluation on the test set.

\subsection{Settings}
In our study, since all types of nodes in the MedGraph have connections to encounters, to avoid over-smoothing issues \cite{li2018deeper}, we set a 1-layer MedGCN with the output dimension 300 and dropout rate 0.1. $f_{\theta_M}$ and $f_{\theta_L}$ are both single layer fully-connected neural networks. The features of P, E, M, L nodes are all initialized with one-hot vectors. We implement MedGCN with PyTorch 1.0.1, and train it using Adam optimizer \cite{kingma2014adam} with learning rate 0.001. The max training epoch is set 300. The cross regularization coefficient $\lambda=1$.

\subsection{Performance for Medication Recommendation}

\subsubsection{Baselines}
For medication recommendation, besides the regular MedGCN, we also implemented the following models for comparison. To release the workload of physicians in diagnostic reasoning, our model was designed not to require diagnostic input. Advanced deep learning models that require the diagnosis code as an input (e.g., GameNet \cite{shang2019gamenet}, G-BERT \cite{shang2019pre} and DMNC \cite{le2018dual}) cannot be directly applied to our scenario, thus not included in our baselines. Since our MedGraph only involves lab nodes and patient nodes besides the encounter nodes and medication nodes, we only consider the models that accept lab tests and patients as input. 
\begin{itemize}
    \item \textbf{MedGCN-ind}: the inductive version of MedGCN where the test encounters and all their connections are removed from MedGraph in the training process.
    \item \textbf{MedGCN-Med}: the MedGCN trained without cross regularization using Eq. \ref{eq:Mloss} as the loss function.
    \item \textbf{Multi-layer perceptron (MLP)}: we implement a 3-layer MLP appending a softmax output layer for multi-label classification. The MLP has a hidden layer size 100 with a Relu activation and an Adam solver \cite{kingma2014adam}. In our case, the input dimension is the sum of lab test counts and patient counts; the output dimension is the number of medications in considering.
    \item \textbf{Gradient Boosting Decision Tree (GBDT)} \cite{friedman2002stochastic}: GBDT is an accurate and effective procedure that can be used for both regression and classification problems in a variety of areas. GBDT produces a prediction model in the form of an ensemble of weak decision trees. In our case, we implement multiple binary GBDT classifiers corresponding to the medications based on lab test values and patients.
    \item \textbf{Random Forest (RF)} \cite{breiman2001random}: RF is a classifier that fits a number of decision tree classifiers on various sub-samples of the dataset and uses averaging to improve the predictive accuracy and control over-fitting. In our case, we implement multiple binary RF classifiers corresponding to the medications based on lab test values and patients. For each RF classifier, we set 100 decision trees in the forest.
    \item \textbf{Logistic Regression (LR)} \cite{fan2008liblinear}: we implement multiple binary LR classifiers corresponding to the medications based on the liblinear solver \cite{fan2008liblinear}. For each LR classifier, the input features consist of lab test values and patients.
    \item \textbf{Support Vector Machine (SVM)} \cite{chang2011libsvm}: we implement multiple binary SVM classifiers corresponding to the medications based on libsvm \cite{chang2011libsvm}. For each SVM classifier, we set the rbf kernel and the input features consist of lab test values and patients.
    \item \textbf{Classifier Chains (CC)} \cite{read2011classifier}: a popular multi-label learning method that models the correlation between labels by feeding both input and previous classification results into the latter classifiers. We leverage SVM as the base estimator.
\end{itemize}
For the baselines that cannot handle missing values (MLP, GBDT, RF, LR, SVM, CC), we simply replaced missing values in lab test with 0. We have also tried replacing missing values with mean values, the results did not show much difference. All the baselines were implemented with scikit-learn \cite{scikit-learn}. Since the dataset is imbalance, all the baselines are implemented with balanced class weight. 

\subsubsection{Metrics}
Since the average number of medications for an encounter is {about} 2 in our datasets {(specifically, 2.0 in NMEDW and 1.3 in MIMIC-III)}, we used mean average precision at 2 (MAP@2) to evaluate the performance of medication recommendation. MAP is a well-known metric to evaluate the performance of a recommender system or information retrieval system\footnote{Refer to \url{http://fastml.com/what-you-wanted-to-know-about-mean-average-precision/} for details.}. Since the task of medication recommendation can be regarded as a multi-label classification problem as well, we also used label ranking average precision (LRAP) \footnote{Refer to \url{https://scikit-learn.org/stable/modules/model_evaluation.html#multilabel-ranking-metrics} for details.} to evaluate the classification performance of all the considered methods. Both MAP and LRAP would be in the range of $[0,1]$, and higher LRAP or MAP indicates more accuracy recommendation. 

\subsubsection{Results}
For all the methods, we executed the training and test processes 5 times with different initializations, and recorded the average performance ($avg.$) and standard deviation ($std.$) in the form of $avg. \pm std.$. The results for medication recommendation are listed in Table \ref{tab:medcine} and \ref{tab:medicine_mimic} for NMEDW and MIMIC-III, respectively. Since LR and SVM are not sensitive to the initialization, the $std.$ are not listed. The results show that all the three variants of MedGCN (i.e., MedGCN, MedGCN-ind and MedGCN-Med) significantly outperform all the baselines in terms of both LRAP and MAP@2 ($p<0.05$ with t-test). MedGCN also significantly outperforms MedGCN-Med in both terms. MedGCN-ind is comparable to MedGCN on both datasets.

Comparing MedGCN-Med and MedGCN, the only difference between the two models is the loss function where MedGCN was regularized by the lab test imputation task, while MedGCN-Med did not, the performance of MedGCN is much better than MedGCN-Med, which validates the efficacy of the cross regularization. We also found that MedGCN-ind is comparable to MedGCN, indicating that MedGCN-ind can handle new coming data very well, which validates the potential of our system for online medication recommendation.  

\begin{table}[htbp]
  \centering
   \caption{Results for medication recommendation {on NMEDW dataset}. The best results are bolded. ``*'' indicates the best results significantly outperform this result ($p<0.05$ with t-test).}
    \begin{tabular}{l|ll}
    \toprule
     Methods & {LRAP} & {MAP@2}   \\
    \midrule
    MedGCN (ours) & \textbf{.7588$\pm$.0028} & \textbf{.7558$\pm$.0035}  \\
    MedGCN-ind (ours) & .7491$\pm$.0067* & .7558$\pm$.0073  \\
    MedGCN-Med (ours) & {.7477$\pm$.0032}* & {.7457$\pm$.0046}* \\
    MLP   & {.7331$\pm$.0126}* & {.6965$\pm$.0113}*  \\ 
    GBDT \cite{friedman2002stochastic}  & {.7120$\pm$.0018*} & {.6864$\pm$.0023*}  \\
    RF \cite{breiman2001random}  & {.6872$\pm$.0072*} & {.7055$\pm$.0068}*  \\ 
    LR \cite{fan2008liblinear}    & {.5325}* & {.4133}*   \\
    SVM \cite{chang2011libsvm}  & {.4324*} & {.3353*}   \\
    {CC \cite{read2011classifier}}   & {.6276$\pm$.0116*} & {.6182$\pm$.0159*} \\
    \bottomrule
    \end{tabular}  \\%
 MedGCN-ind: inductive MedGCN; MedGCN-Med: MedGCN without cross regularization and only use Eq. \ref{eq:Mloss} as loss function; MLP: Multi-layer perceptron; GBDT: Gradient Boosting Decision Tree; RF: Random Forest; LR: Logistic Regression; SVM: Support Vector Machine.
  \label{tab:medcine}%
\end{table}%

\begin{table}[htbp]
  \centering
  \caption{{Results for medication recommendation on MIMIC dataset. The best results are bolded. ``*'' indicates the best results significantly outperform this result ($p<0.05$ with t-test).}}
    \begin{tabular}{l|l|l}
    \toprule
    Methods & {LRAP} & {MAP@2} \\
    \midrule
    MedGCN (ours) & \textbf{.8349$\pm$.0008} & {.8069$\pm$.0022} \\
    MedGCN-ind (ours) & {.8345$\pm$.0007} & \textbf{.8070$\pm$.0029} \\
    MedGCN-Med (ours) & {.8346$\pm$.0005} & {.8061$\pm$.0020} \\
    MLP   & {.8325$\pm$.0003}* & {.8030$\pm$.0030}* \\
    GBDT \cite{friedman2002stochastic} & {.5793$\pm$.0001}* & {.5019$\pm$.0002}* \\
    RF \cite{breiman2001random}    & {.8215$\pm$.0007}* & {.8030$\pm$.0011}* \\
    LR \cite{fan2008liblinear}    & .3367* & .1839* \\
    SVM \cite{chang2011libsvm}  & .6642* & .6146* \\
    CC \cite{read2011classifier}    & .7660$\pm$.0005* & .7153$\pm$.0003* \\
     \bottomrule
    \end{tabular}%
  \label{tab:medicine_mimic}%
\end{table}%

\subsection{Performance for Lab Test Imputation}

\subsubsection{Baselines}
For lab test imputation, besides the regular MedGCN, we also implemented the following models for comparison.
\begin{itemize}
    \item \textbf{MedGCN-ind}: the inductive version of MedGCN where the test encounters and all their connections are removed from MedGraph in the training process.
    \item \textbf{MedGCN-Lab}: MedGCN-Lab is the MedGCN trained without cross regularization using Eq. \ref{eq:Lloss} as loss function.
    \item \textbf{Multivariate Imputation by Chained Equations (MICE)} \cite{buuren2010mice}: MICE is a popular multiple imputation method used to replace missing data values based on fully conditional specification, where each incomplete variable is imputed by a separate model. In our case, we directly use MICE to impute the adjacency matrix $A_{E\times L}$.
    \item \textbf{Multi-Graph Convolutional Neural Networks (MGCNN)} \cite{monti2017geometric}: MGCNN is a  geometric deep learning method on graphs for matrix completion by combining a multi-graph convolutional neural network and a recurrent neural network. We applied MGCNN to complete the adjacency matrix $A_{E\times L}$ to implement the imputation. 
    \item \textbf{Graph Convolutional Matrix Completion (GCMC)} \cite{berg2018graph}: GCMC uses a Graph convolutional encoder and a bilinear decoder to predict the connections among nodes in a bipartite graph, thus, to complete adjacency matrix of the bipartite graph. We applied GCMC to complete the adjacency matrix $A_{E\times L}$ to implement the imputation.
    \item \textbf{GCMC+FEAT} \cite{berg2018graph}: GCMC with medication information as side information for encounters. 
\end{itemize}
Since MGCNN, GCMC and GCMC+FEAT are designed for discrete features, when testing these baselines, we discretized the continuous lab value into 5 ratings for each lab in advance.

{\subsubsection{Metrics}}
We use mean square error (MSE) to measure the performance of lab test imputation, following previous studies \cite{luo2016using}. A lower MSE indicates the predicted values are more close to the true value, thus a better model.

{\subsubsection{Results}}
We also repeated the training and test processes 5 times with different initializations for all the methods, and recorded the results of lab test imputation in the form of $avg. \pm std.$ in Table \ref{tab:lab} and \ref{tab:lab_mimic} for NMEDW and MIMIC-III, respectively. The results show MedGCN can significantly perform better than all other methods ($p<0.05$ with t-test), and MedGCN-ind and MedGCN-Lab both perform better than the other baselines. Comparing the performance of MedGCN-Lab and MedGCN, we again validate the efficacy of the cross regularization.  Although MedGCN-ind is not comparable to MedGCN and MedGCN-Lab {on NMEDW} for lab test imputation, it can outperform the other baselines and provide acceptable results. 

\begin{table}[htbp]
  \centering
  \caption{Results for lab test imputation {on NMEDW dataset}. The best results are bolded. ``*'' indicates the best results significantly outperform this result ($p<0.05$ with t-test). }
    \begin{tabular}{l|c}
    \toprule
     Methods &  MSE \\
    \midrule
     MedGCN (ours) &  \textbf{.0229$\pm$.0025} \\
     MedGCN-ind (ours) & .0264$\pm$.0034* \\
     MedGCN-Lab (ours) & .0254$\pm$.0003* \\
    { MICE \cite{buuren2010mice}} &  {.0474$\pm$.0010*   } \\
    MGCNN \cite{monti2017geometric} &        .0369$\pm$.0009* \\
    GCMC \cite{berg2018graph} &       .0426$\pm$.0025* \\
    GCMC+FEAT \cite{berg2018graph}&    .0359$\pm$.0030* \\
    \bottomrule
    \end{tabular} \\%
 MedGCN-Lab: MedGCN without cross regularization and only use Eq. \ref{eq:Lloss} as loss function; MGCNN: Multi-Graph Convolutional Neural Networks; GCMC: Graph Convolutional Matrix Completion; GCMC+FEAT: GCMC with side information
  \label{tab:lab}%
\end{table}%

\begin{table}[htbp]
  \centering
  \caption{{Results for lab test imputation on MIMIC-III dataset. The best results are bolded. ``*'' indicates the best results significantly outperform this result ($p<0.05$ with t-test). } }
    \begin{tabular}{l|c}
    \toprule
    Methods & MSE \\
    \midrule
    MedGCN(ours) & \textbf{.0140$\pm$.0002} \\
    MedGCN-ind(ours) & .0143$\pm$.0002* \\
    MedGCN-Lab(ours) & .0143$\pm$.0001* \\
    MICE \cite{buuren2010mice}  & .0146$\pm$.0001* \\
    MGCNN \cite{monti2017geometric} &        .0413$\pm$.0048*\\
    GCMC \cite{berg2018graph} & .0296$\pm$.0004* \\
    GCMC+FEAT \cite{berg2018graph} & .0290$\pm$.0001* \\
    \bottomrule
    \end{tabular}%
  \label{tab:lab_mimic}%
\end{table}%

To validate the efficacy of MedGCN for lab test imputation, we also test our imputation results of lab tests for medication recommendation. We compare the results of our imputation with lab values filled with 0 in Table \ref{tab:MR_lab_fill}. We found that using labs filled by MedGCN can achieve better performance than that filled with 0 for most of the classifiers. Only for the GBDT classifier, filled with 0 is better than filled by MedGCN, but the differences are not significant. 

\begin{table}[htbp]
  \centering
  \caption{{Results of Medication recommendation on NMEDW dataset using lab tests filled by MedGCN and with 0. }}
    \begin{tabular}{l|c|c|c|c}
    \toprule
    {\multirow{2}[0]{*}{methods}} & \multicolumn{2}{c|}{Filled with 0} & \multicolumn{2}{c}{Filled by MedGCN} \\
    \cmidrule{2-3} \cmidrule{4-5}
          & {LRAP} & {MAP@2} & LRAP  & MAP@2 \\
          \midrule
    MLP   & {.7331$\pm$.0126} & {.6965$\pm$.0113} & \textbf{.7409$\pm$.0040} & \textbf{.7448$\pm$.0165} \\
    GBDT  & \textbf{{.7120$\pm$.0018}} & \textbf{{.6864$\pm$.0023}} & .6968$\pm$.0105 & .6832$\pm$.0215 \\
    RF    & {.6872$\pm$.0072} & {.7055$\pm$.0068} & \textbf{.6907$\pm$.0052} & \textbf{.7303$\pm$.00167} \\
    LR    & 0.5325 & 0.4113 & \textbf{0.5337} & \textbf{0.4408} \\
    SVM   & 0.4324 & 0.3353 & \textbf{0.5553} & \textbf{0.5376} \\
    \bottomrule
    \end{tabular}%
  \label{tab:MR_lab_fill}%
\end{table}%

\subsection{Parameter Sensitivity}
\begin{figure}[!t]
  \centering
   \subfloat[LRAP]{\includegraphics[width=.5\columnwidth,page=1]{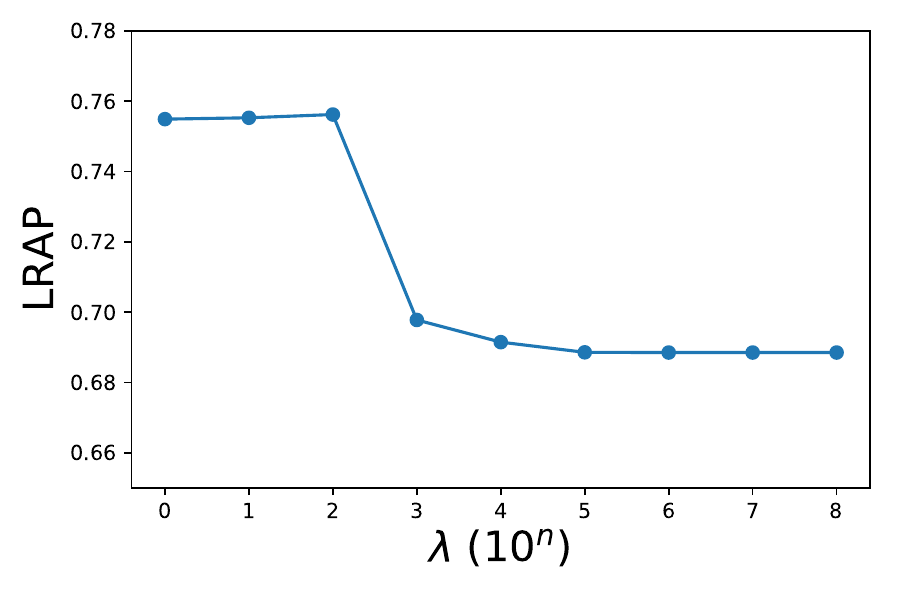}\label{fig:lambda-lrap}}
   \subfloat[MAP@2]{\includegraphics[width=.5\columnwidth,page=2]{lambda_MR.pdf}\label{fig:lambda-map}}
     \caption{{Performances of MedGCN with different cross regularization parameters $\lambda$ for the medication recommendation. x axis is exponential. }}
     \label{fig:lambda-med}
\end{figure}

\begin{figure}[!t]   
\centering
\includegraphics[width=.5\columnwidth,page=3]{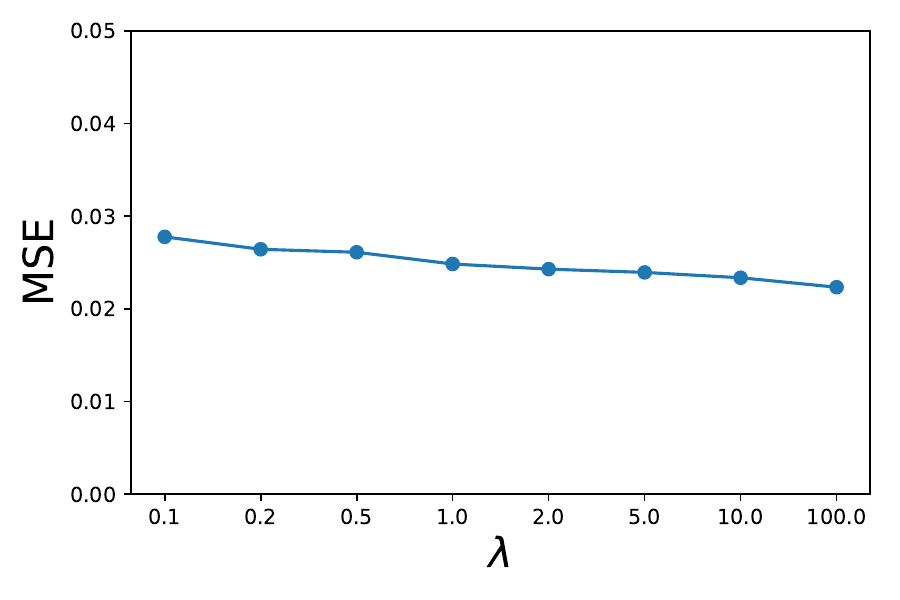}
  \caption{MSE performances of MedGCN with different cross regularization parameters $\lambda$ for lab test imputation. }
  \label{fig:lambda-lab}
\end{figure}

We have also tested the effect of cross regularization parameter $\lambda$ of MedGCN on the results for the two tasks {on NMEDW dataset} (see Section \ref{sec:training} for the detail of cross regularization). Figure \ref{fig:lambda-med} shows the performances of MedGCN for medication recommendation in terms of LRAP (Figure \ref{fig:lambda-lrap}) and MAP@2 (Figure \ref{fig:lambda-map}) with different $\lambda$. Figure \ref{fig:lambda-lab} shows the MSEs of MedGCN for lab test imputation with different $\lambda$. From the variation curves of LRAP (Figure \ref{fig:lambda-lrap}), MAP@2 (Figure \ref{fig:lambda-map}) , and MSE (Figure \ref{fig:lambda-lab}), all the 3 metrics change very slowly with the hyper-parameter $\lambda$ {(note that the x tick in Figure \ref{fig:lambda-lrap} and \ref{fig:lambda-map} is exponential)}, indicating that MedGCN is robust to $\lambda$ for both medication recommendation and lab test imputation. {From Figure \ref{fig:lambda-lrap} and \ref{fig:lambda-map}, the performance of medication recommendation decreases with $\lambda$ increases exponentially. From Figure \ref{fig:lambda-lab}, the performance of lab test imputation increases slowly with $\lambda$ increases. } Because $\lambda$ is to adjust the weights of the two learning tasks, a larger $\lambda$ represents the model is more prone to the lab test imputation task, and vice versa, thus, MedGCN with a greater $\lambda$ will have a better lab test imputation performance but a slightly lower medication recommendation performance.

\begin{figure}[!t]
  \centering
   \subfloat[medication recommendation]{\includegraphics[width=.5\columnwidth,page=1]{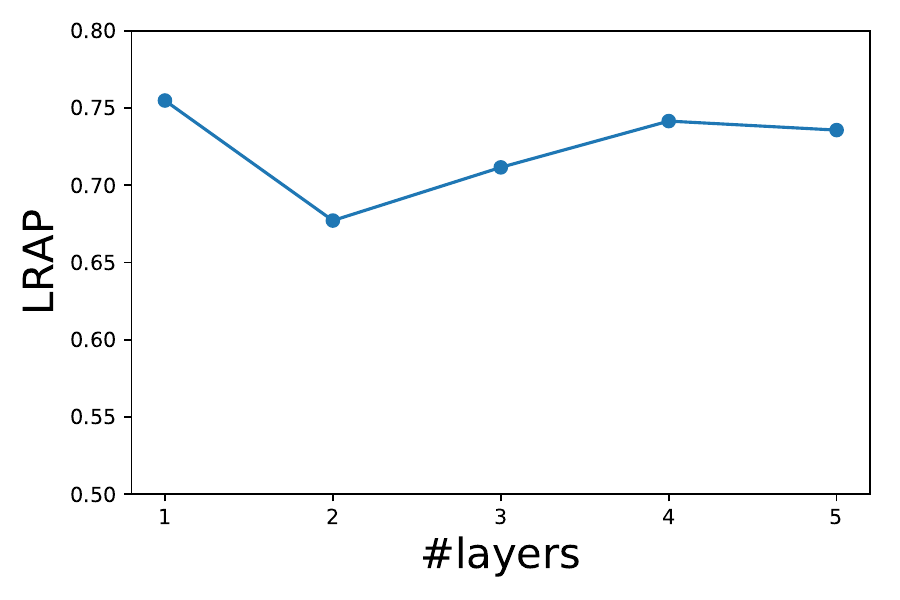}}
   \subfloat[lab test imputation]{\includegraphics[width=.5\columnwidth,page=2]{layer.pdf}}
  \caption{{Performance of MedGCN with different number of layers for the two tasks. } }
  \label{fig:layer}
\end{figure}

We have also tested multi-layer MedGCNs for the two tasks, the performance of MedGCN with different number of layers are shown in Figure \ref{fig:layer}, each layer has 300 units in the experiment. From Figure \ref{fig:layer}, the 1-layer MedGCN performs better than multi-layer MedGCNs for both medication recommendation and lab test imputation. Because in the constructed MedGraph, an encounter and its one-hop neighbors are much more important than other neighbors for the two learning tasks, stacking multiple MedGCN layers would result in over-smoothing issues \cite{li2018deeper}.

\subsection{Case Study}
In our case study, we choose an encounter from NMEDW that has only a few lab values available from the test set to see what medication our model will recommend, and what lab values will be imputed. The results of our case study are shown in Table \ref{tab:casestudy}, where the upper part and the middle part respectively list the recommended medications and estimated lab values based on two available labs on urine analysis, and the lower part lists the abbreviations of appeared medications and labs. Of note, albuminuria was shown to associate with a 2.4-fold likelihood of risk of lung cancer~\cite{jorgensen2008association}. 

In the upper part of Table \ref{tab:casestudy}, the true medications prescribed by physicians (ground truth) and the top 5 recommended medications by different methods are listed. Although no priority exists in the ground truth, for the implemented recommendation models, medications in the front of the recommendation list have higher recommendation priority. We found the top 3 recommendations of MedGCN exactly match the physician prescribed medications. LR and MLP missed one medication in the top 5 recommendations; GBDT recommends a lower priority for \textit{gem} than two other medications (i.e., \textit{car} and \textit{bev}) that physicians did not prescribe. 

The middle part of Table \ref{tab:casestudy} lists the imputed values for lab test \textit{urobi} for the same encounter, the true value was measured as 0.2, but we removed it in the training process. We can see that our MedGCN predicts (0.19) closer to the true value than all other methods. What is more, MedGCN can perform the two tasks simultaneously.

\begin{table}[!t]
  \centering
  \caption{The recommended medications and estimated lab values in our case study with two labs available. }
    \begin{tabular}{l|l|l}
    \toprule
    {\textbf{Available labs}} & \textbf{Methods} & \textbf{Recommendations} \\
    \midrule
    {\textit{uph}: 6} & Ground Truth & {\textit{dex},	\textit{gem},	\textit{ond}		} \\
    {\textit{uspg}: 1.019} & MedGCN & {\textit{ond},	\textit{dex},	\textit{gem},	\textit{peg},	\textit{bev}} \\
          & LR    & {\textit{ond},	\textit{dex},	\textit{pem},	\textit{bev},	\textit{car}} \\
          & MLP   & {\textit{ond},	\textit{dex},	\textit{hyd},	\textit{ral},	\textit{den}} \\
          & GBDT  & {\textit{ond},	\textit{dex},	\textit{car},	\textit{bev},	\textit{gem}} \\
    \midrule   
    \midrule
     \textbf{Available labs}  & \textbf{Methods} & \textbf{estimated lab} (\textit{urobi}), EU/dL \\
    \midrule       
      {\textit{uph}: 6} & True value & 0.2 \\
      {\textit{uspg}: 1.019} & {MedGCN} & 0.19 \\
            & GCMC &  0.25 \\
              & GCMA+FEAT &  0.24  \\
               & MGCNN &  1.01  \\
    \midrule
    \midrule
    lab abbreviations & \textit{uph} & {Urine pH} \\
          & \textit{uspg} & {Urine specific gravity} \\
          & \textit{urobi} & {Urobilinogen} \\
    \midrule
    med abbreviations & \textit{ond} & {ondansetron} \\
                     & \textit{dex} & {dexamethasone} \\
                     & \textit{gem} & {gemcitabine} \\
                     & \textit{peg} & {pegfilgrastim} \\
                     & \textit{bev} & {bevacizumab} \\
                     & \textit{pem} & {pemetrexed} \\
                     & \textit{car} & {carboplatin} \\
                     & \textit{hyd} & {hydroxyurea} \\
                     & \textit{ral} & {raloxifene} \\
                     & \textit{den} & {denosumab} \\
          \bottomrule
    \end{tabular}%
  
  \label{tab:casestudy}%
\end{table}%

\subsection{Discussion}
From the above results, MedGCN works well for both medication recommendation and lab test imputation.
The reason may be two-fold. First, because the constructed MedGraph incorporates the complex relations between different medical entities, MedGCN built on the informative MedGraph has the potential to learn a very informative representation for each node. Considering the baselines in Tables \ref{tab:medcine} and \ref{tab:lab}, they all ignore the correlations between different medications as well as different labs, while MedGCN takes these relations into account through their shared encounters in the constructed MedGraph, thus MedGCN can achieve better performances for both tasks. Furthermore, the cross regularization strategy enhances MedGCN by reducing overfitting. MedGCN uses a loss function considering multiple tasks to guide the training of the model by the cross regularization strategy, thus reducing the overfitting for a specific task and making the learned representation predictive for multiple tasks. The results in Tables \ref{tab:medcine} and \ref{tab:lab} also validate the efficacy of cross regularization. 

Besides the improved performance, MedGCN also has some other features. (1) MedGCN can incorporate the features of quite different medical tasks in one model and perform multiple tasks simultaneously. MedGCN takes the advantages of multi-task learning \cite{ruder2017overview,zhang2017survey} and incorporates multiple medical tasks such as medication recommendation and lab test imputation in one single model. In our future work, more medical tasks will be incorporated into MedGCN. (2) MedGCN can handle missing attribute values. Most previous GCN models cannot accept missing values. We convert the node attributes as a new type of nodes incorporated to MedGraph where a missing value corresponds to a missing edge that GCN can handle well. (3) MedGCN can be applied to heterogeneous graphs. MedGraph is a typical heterogeneous graph that incorporates many types of medical entities and relations. We split the heterogeneous graph into multiple subgraphs with each subgraph has only one type of edge. In each GCN layer, the representations of each node in all the subgraphs are aggregated as the node representation. (4) MedGCN enables inductive learning on MedGraphs. Tables \ref{tab:medcine} and \ref{tab:lab} show the inductive version of MedGCN almost performs comparable to regular MedGCN, validating the generalization ability of our model to new data.

Another feature of our work is that we do not involve diagnosis code in our framework for medication recommendation and lab test imputation. This could release physicians from heavy lifting in diagnostic reasoning. To recommend medications for a new patient/encounter, unlike most of the previous work that needs physicians to provide the diagnosis codes \cite{shang2019gamenet, zhang2017leap, jin2020syndrome, le2018dual, shang2019pre}, our model just needs the results of only a few lab tests.

In this paper, MedGCN is evaluated with a simple Medgraph with 4 types of medical entities and two medical tasks. In practical scenarios, a MedGraph can be scaled to include more medical entities as well as the relations, such as diagnosis code, and the MedGCN trained on the MedGraph can be used to perform other tasks such as diagnosis prediction. Also, the time order of encounters can be considered to construct a directional MedGraph to enhance the model. These represent the direction of our future work.

\section{Conclusion}
In this article, we innovatively incorporated the complex associations between multiple medical entities into MedGraph, and developed MedGCN, an end-to-end graph embedding model, to learn informative medical entity representations for multiple medical tasks based on MedGraph. We augmented MedGCN to handle heterogeneous MedGraph by formulating different medical entities different types of nodes. By transforming node features to a new type of nodes in the graph, MedGCN can also handle missing values in the features of a medical entity. MedGCN is a general inductive model that could use the learned node representations and network weights to efficiently generate representations for new nodes. In addition, we introduce cross regularization to reduce overfittings by making the learned representation more informative. Experimental results on a real medical dataset showed that MedGCN outperformed state-of-the-art methods in both medication recommendation and lab test imputation tasks.


\section*{Acknowledgements}

The research is supported in part by the following US NIH grants: R21LM012618, 5UL1TR001422, U01TR003528  and  R01LM013337.



\bibliographystyle{unsrt}
\bibliography{medgcn}

\begin{thebibliography}{10}

\bibitem{winslow2012computational}
Raimond~L Winslow, Natalia Trayanova, Donald Geman, and Michael~I Miller.
\newblock Computational medicine: translating models to clinical care.
\newblock {\em Science translational medicine}, 4(158):158rv11--158rv11, 2012.

\bibitem{karczewski2012pharmacogenomics}
Konrad~J Karczewski, Roxana Daneshjou, and Russ~B Altman.
\newblock Pharmacogenomics.
\newblock {\em PLoS computational biology}, 8(12):e1002817, 2012.

\bibitem{kohane2015ten}
Isaac~S Kohane.
\newblock Ten things we have to do to achieve precision medicine.
\newblock {\em Science}, 349(6243):37--38, 2015.

\bibitem{shang2019gamenet}
Junyuan Shang, Cao Xiao, Tengfei Ma, Hongyan Li, and Jimeng Sun.
\newblock Gamenet: graph augmented memory networks for recommending medication
  combination.
\newblock In {\em Proceedings of the AAAI Conference on Artificial
  Intelligence}, volume~33, pages 1126--1133, 2019.

\bibitem{wang2019neural}
Jonathan~X Wang, Delaney~K Sullivan, Adam~J Wells, Alex~C Wells, and Jonathan~H
  Chen.
\newblock Neural networks for clinical order decision support.
\newblock {\em AMIA Summits on Translational Science Proceedings}, 2019:315,
  2019.

\bibitem{zhang2017leap}
Yutao Zhang, Robert Chen, Jie Tang, Walter~F Stewart, and Jimeng Sun.
\newblock Leap: learning to prescribe effective and safe treatment combinations
  for multimorbidity.
\newblock In {\em proceedings of the 23rd ACM SIGKDD international conference
  on knowledge Discovery and data Mining}, pages 1315--1324. ACM, 2017.

\bibitem{yu2020predict}
Lishan Yu, Qiuchen Zhang, Elmer~V Bernstam, and Xiaoqian Jiang.
\newblock Predict or draw blood: An integrated method to reduce lab tests.
\newblock {\em Journal of Biomedical Informatics}, page 103394, 2020.

\bibitem{waljee2013comparison}
Akbar~K Waljee, Ashin Mukherjee, Amit~G Singal, Yiwei Zhang, Jeffrey Warren,
  Ulysses Balis, Jorge Marrero, Ji~Zhu, and Peter~DR Higgins.
\newblock Comparison of imputation methods for missing laboratory data in
  medicine.
\newblock {\em BMJ open}, 3(8):e002847, 2013.

\bibitem{luo2016using}
Yuan Luo, Peter Szolovits, Anand~S Dighe, and Jason~M Baron.
\newblock Using machine learning to predict laboratory test results.
\newblock {\em American journal of clinical pathology}, 145(6):778--788, 2016.

\bibitem{luo20173d}
Yuan Luo, Peter Szolovits, Anand~S Dighe, and Jason~M Baron.
\newblock 3d-mice: integration of cross-sectional and longitudinal imputation
  for multi-analyte longitudinal clinical data.
\newblock {\em Journal of the American Medical Informatics Association},
  25(6):645--653, 2017.

\bibitem{wu2020comprehensive}
Zonghan Wu, Shirui Pan, Fengwen Chen, Guodong Long, Chengqi Zhang, and S~Yu
  Philip.
\newblock A comprehensive survey on graph neural networks.
\newblock {\em IEEE Transactions on Neural Networks and Learning Systems},
  2020.

\bibitem{kipf2017semi}
Thomas~N Kipf and Max Welling.
\newblock Semi-supervised classification with graph convolutional networks.
\newblock In {\em International Conference on Learning Representations}, 2017.

\bibitem{hamilton2017inductive}
Will Hamilton, Zhitao Ying, and Jure Leskovec.
\newblock Inductive representation learning on large graphs.
\newblock In {\em Advances in neural information processing systems}, pages
  1024--1034, 2017.

\bibitem{chen2018fastgcn}
Jie Chen, Tengfei Ma, and Cao Xiao.
\newblock Fast{GCN}: Fast learning with graph convolutional networks via
  importance sampling.
\newblock In {\em International Conference on Learning Representations}, 2018.

\bibitem{li2018classifying}
Yifu Li, Ran Jin, and Yuan Luo.
\newblock Classifying relations in clinical narratives using segment graph
  convolutional and recurrent neural networks (seg-gcrns).
\newblock {\em Journal of the American Medical Informatics Association},
  26(3):262--268, 2018.

\bibitem{yao2019graph}
Liang Yao, Chengsheng Mao, and Yuan Luo.
\newblock Graph convolutional networks for text classification.
\newblock In {\em Proceedings of the AAAI Conference on Artificial
  Intelligence}, volume~33, pages 7370--7377, 2019.

\bibitem{mao2019imagegcn}
Chengsheng Mao, Liang Yao, and Yuan Luo.
\newblock Imagegcn: Multi-relational image graph convolutional networks for
  disease identification with chest x-rays.
\newblock {\em arXiv preprint arXiv:1904.00325}, 2019.

\bibitem{tran2015learning}
Truyen Tran, Tu~Dinh Nguyen, Dinh Phung, and Svetha Venkatesh.
\newblock Learning vector representation of medical objects via emr-driven
  nonnegative restricted boltzmann machines (enrbm).
\newblock {\em Journal of biomedical informatics}, 54:96--105, 2015.

\bibitem{miotto2016deep}
Riccardo Miotto, Li~Li, Brian~A Kidd, and Joel~T Dudley.
\newblock Deep patient: an unsupervised representation to predict the future of
  patients from the electronic health records.
\newblock {\em Scientific reports}, 6(1):1--10, 2016.

\bibitem{sun2020disease}
Zhenchao Sun, Hongzhi Yin, Hongxu Chen, Tong Chen, Lizhen Cui, and Fan Yang.
\newblock Disease prediction via graph neural networks.
\newblock {\em IEEE Journal of Biomedical and Health Informatics},
  25(3):818--826, 2020.

\bibitem{choi2016doctor}
Edward Choi, Mohammad~Taha Bahadori, Andy Schuetz, Walter~F Stewart, and Jimeng
  Sun.
\newblock Doctor ai: Predicting clinical events via recurrent neural networks.
\newblock In {\em Machine learning for healthcare conference}, pages 301--318.
  PMLR, 2016.

\bibitem{choi2016multi}
Edward Choi, Mohammad~Taha Bahadori, Elizabeth Searles, Catherine Coffey,
  Michael Thompson, James Bost, Javier Tejedor-Sojo, and Jimeng Sun.
\newblock Multi-layer representation learning for medical concepts.
\newblock In {\em proceedings of the 22nd ACM SIGKDD international conference
  on knowledge discovery and data mining}, pages 1495--1504, 2016.

\bibitem{lee2020harmonized}
Dongha Lee, Xiaoqian Jiang, and Hwanjo Yu.
\newblock Harmonized representation learning on dynamic ehr graphs.
\newblock {\em Journal of biomedical informatics}, 106:103426, 2020.

\bibitem{zitnik2018modeling}
Marinka Zitnik, Monica Agrawal, and Jure Leskovec.
\newblock Modeling polypharmacy side effects with graph convolutional networks.
\newblock {\em Bioinformatics}, 34(13):i457--i466, 2018.

\bibitem{liu2020heterogeneous}
Zheng Liu, Xiaohan Li, Hao Peng, Lifang He, and S~Yu Philip.
\newblock Heterogeneous similarity graph neural network on electronic health
  records.
\newblock In {\em 2020 IEEE International Conference on Big Data (Big Data)},
  pages 1196--1205. IEEE, 2020.

\bibitem{yella2021mgatrx}
Jaswanth~Kumar Yella and Anil Jegga.
\newblock Mgatrx: discovering drug repositioning candidates using multi-view
  graph attention.
\newblock {\em IEEE/ACM Transactions on Computational Biology and
  Bioinformatics}, 2021.

\bibitem{jin2020syndrome}
Yuanyuan Jin, Wei Zhang, Xiangnan He, Xinyu Wang, and Xiaoling Wang.
\newblock Syndrome-aware herb recommendation with multi-graph convolution
  network.
\newblock In {\em 2020 IEEE 36th International Conference on Data Engineering
  (ICDE)}, pages 145--156. IEEE, 2020.

\bibitem{le2018dual}
Hung Le, Truyen Tran, and Svetha Venkatesh.
\newblock Dual memory neural computer for asynchronous two-view sequential
  learning.
\newblock In {\em Proceedings of the 24th ACM SIGKDD International Conference
  on Knowledge Discovery \& Data Mining}, pages 1637--1645. ACM, 2018.

\bibitem{shang2019pre}
Junyuan Shang, Tengfei Ma, Cao Xiao, and Jimeng Sun.
\newblock Pre-training of graph augmented transformers for medication
  recommendation.
\newblock In {\em Proceedings of the Twenty-Eighth International Joint
  Conference on Artificial Intelligence, {IJCAI-19}}, pages 5953--5959.
  International Joint Conferences on Artificial Intelligence Organization, 7
  2019.

\bibitem{buuren2010mice}
S~van Buuren and Karin Groothuis-Oudshoorn.
\newblock mice: Multivariate imputation by chained equations in r.
\newblock {\em Journal of statistical software}, pages 1--68, 2010.

\bibitem{stekhoven2011missforest}
Daniel~J Stekhoven and Peter B{\"u}hlmann.
\newblock Missforest—non-parametric missing value imputation for mixed-type
  data.
\newblock {\em Bioinformatics}, 28(1):112--118, 2011.

\bibitem{he2011functional}
Yulei He, Recai Yucel, and Trivellore~E Raghunathan.
\newblock A functional multiple imputation approach to incomplete longitudinal
  data.
\newblock {\em Statistics in medicine}, 30(10):1137--1156, 2011.

\bibitem{kliethermes2014bayesian}
Stephanie Kliethermes and Jacob Oleson.
\newblock A bayesian approach to functional mixed-effects modeling for
  longitudinal data with binomial outcomes.
\newblock {\em Statistics in medicine}, 33(18):3130--3146, 2014.

\bibitem{zhang2018multi}
Xi~Zhang, Lifang He, Kun Chen, Yuan Luo, Jiayu Zhou, and Fei Wang.
\newblock Multi-view graph convolutional network and its applications on
  neuroimage analysis for parkinson’s disease.
\newblock In {\em AMIA Annual Symposium Proceedings}, volume 2018, page 1147.
  American Medical Informatics Association, 2018.

\bibitem{hsieh2020drug}
Kang-Lin Hsieh, Yinyin Wang, Luyao Chen, Zhongming Zhao, Sean Savitz, Xiaoqian
  Jiang, Jing Tang, and Yejin Kim.
\newblock Drug repurposing for covid-19 using graph neural network with
  genetic, mechanistic, and epidemiological validation.
\newblock {\em Research Square}, 2020.

\bibitem{li2018deeper}
Qimai Li, Zhichao Han, and Xiao-Ming Wu.
\newblock Deeper insights into graph convolutional networks for semi-supervised
  learning.
\newblock In {\em Thirty-Second AAAI Conference on Artificial Intelligence},
  2018.

\bibitem{morris2019weisfeiler}
Christopher Morris, Martin Ritzert, Matthias Fey, William~L Hamilton, Jan~Eric
  Lenssen, Gaurav Rattan, and Martin Grohe.
\newblock Weisfeiler and leman go neural: Higher-order graph neural networks.
\newblock In {\em Proceedings of the AAAI Conference on Artificial
  Intelligence}, volume~33, pages 4602--4609, 2019.

\bibitem{hamilton2017representation}
William~L Hamilton, Rex Ying, and Jure Leskovec.
\newblock Representation learning on graphs: Methods and applications.
\newblock {\em IEEE Data Engineering Bulletin}, 40(3):52--74, 2017.

\bibitem{schlichtkrull2018modeling}
Michael Schlichtkrull, Thomas~N Kipf, Peter Bloem, Rianne Van Den~Berg, Ivan
  Titov, and Max Welling.
\newblock Modeling relational data with graph convolutional networks.
\newblock In {\em European Semantic Web Conference}, pages 593--607. Springer,
  2018.

\bibitem{wang2019heterogeneous}
Xiao Wang, Houye Ji, Chuan Shi, Bai Wang, Yanfang Ye, Peng Cui, and Philip~S
  Yu.
\newblock Heterogeneous graph attention network.
\newblock In {\em The World Wide Web Conference}, pages 2022--2032, 2019.

\bibitem{mao2018deep}
Chengsheng Mao, Liang Yao, Yiheng Pan, Yuan Luo, and Zexian Zeng.
\newblock Deep generative classifiers for thoracic disease diagnosis with chest
  x-ray images.
\newblock In {\em 2018 IEEE International Conference on Bioinformatics and
  Biomedicine (BIBM)}, pages 1209--1214. IEEE, 2018.

\bibitem{wang2017chestxray}
Xiaosong Wang, Yifan Peng, Le~Lu, Zhiyong Lu, Mohammadhadi Bagheri, and Ronald
  Summers.
\newblock Chestx-ray8: Hospital-scale chest x-ray database and benchmarks on
  weakly-supervised classification and localization of common thorax diseases.
\newblock In {\em 2017 IEEE Conference on Computer Vision and Pattern
  Recognition(CVPR)}, pages 3462--3471, 2017.

\bibitem{starren2015enabling}
Justin~B Starren, Andrew~Q Winter, and Donald~M Lloyd-Jones.
\newblock Enabling a learning health system through a unified enterprise data
  warehouse: the experience of the northwestern university clinical and
  translational sciences (nucats) institute.
\newblock {\em Clinical and translational science}, 8(4):269, 2015.

\bibitem{johnson2016mimic}
Alistair~EW Johnson, Tom~J Pollard, Lu~Shen, H~Lehman Li-Wei, Mengling Feng,
  Mohammad Ghassemi, Benjamin Moody, Peter Szolovits, Leo~Anthony Celi, and
  Roger~G Mark.
\newblock Mimic-iii, a freely accessible critical care database.
\newblock {\em Scientific data}, 3(1):1--9, 2016.

\bibitem{kingma2014adam}
Diederik~P Kingma and Jimmy Ba.
\newblock Adam: A method for stochastic optimization.
\newblock In {\em International Conference on Learning Representations}, 2015.

\bibitem{friedman2002stochastic}
Jerome~H Friedman.
\newblock Stochastic gradient boosting.
\newblock {\em Computational statistics \& data analysis}, 38(4):367--378,
  2002.

\bibitem{breiman2001random}
Leo Breiman.
\newblock Random forests.
\newblock {\em Machine learning}, 45(1):5--32, 2001.

\bibitem{fan2008liblinear}
Rong-En Fan, Kai-Wei Chang, Cho-Jui Hsieh, Xiang-Rui Wang, and Chih-Jen Lin.
\newblock Liblinear: A library for large linear classification.
\newblock {\em the Journal of machine Learning research}, 9:1871--1874, 2008.

\bibitem{chang2011libsvm}
Chih-Chung Chang and Chih-Jen Lin.
\newblock Libsvm: a library for support vector machines.
\newblock {\em ACM transactions on intelligent systems and technology (TIST)},
  2(3):27, 2011.

\bibitem{read2011classifier}
Jesse Read, Bernhard Pfahringer, Geoff Holmes, and Eibe Frank.
\newblock Classifier chains for multi-label classification.
\newblock {\em Machine learning}, 85(3):333--359, 2011.

\bibitem{scikit-learn}
F.~Pedregosa, G.~Varoquaux, A.~Gramfort, V.~Michel, B.~Thirion, O.~Grisel,
  M.~Blondel, P.~Prettenhofer, R.~Weiss, V.~Dubourg, J.~Vanderplas, A.~Passos,
  D.~Cournapeau, M.~Brucher, M.~Perrot, and E.~Duchesnay.
\newblock Scikit-learn: Machine learning in {P}ython.
\newblock {\em the Journal of machine Learning Research}, 12:2825--2830, 2011.

\bibitem{monti2017geometric}
Federico Monti, Michael Bronstein, and Xavier Bresson.
\newblock Geometric matrix completion with recurrent multi-graph neural
  networks.
\newblock In {\em Advances in neural information processing systems}, pages
  3697--3707, 2017.

\bibitem{berg2018graph}
Rianne van~den Berg, Thomas~N Kipf, and Max Welling.
\newblock Graph convolutional matrix completion.
\newblock In {\em SIGKDD, Deep Learning Day}, 2018.

\bibitem{jorgensen2008association}
Lone J{\o}rgensen, Ivar Heuch, Trond Jenssen, and Bjarne~K Jacobsen.
\newblock Association of albuminuria and cancer incidence.
\newblock {\em Journal of the American Society of Nephrology}, 19(5):992--998,
  2008.

\bibitem{ruder2017overview}
Sebastian Ruder.
\newblock An overview of multi-task learning in deep neural networks.
\newblock {\em arXiv preprint arXiv:1706.05098}, 2017.

\bibitem{zhang2017survey}
Yu~Zhang and Qiang Yang.
\newblock A survey on multi-task learning.
\newblock {\em arXiv preprint arXiv:1707.08114}, 2017.

\end{thebibliography}

\end{document}